%% file: acl_latex.tex
\documentclass[11pt]{article}
\usepackage{enumitem}
\usepackage{courier}  
\usepackage{booktabs}       
\usepackage{helvet}  
\usepackage{amssymb}
\usepackage{xcolor}

\usepackage[preprint]{acl}

\usepackage{times}
\usepackage{latexsym}
\usepackage{hyperref}
\usepackage{array}

\usepackage[T1]{fontenc}

\usepackage[utf8]{inputenc}

\usepackage{microtype}

\usepackage{inconsolata}

\usepackage{graphicx}

\title{On the Retention of Edited Knowledge in Fine-tuned Language Models}

\author{Fufang Wen \\
  Columbia University \\
  \texttt{fw2325@columbia.edu} \\\And
  Shichang Zhang\thanks{Corresponding author} \\
  Harvard University \\
  \texttt{shzhang@hbs.edu} \\
  }

\begin{document}
\maketitle
\begin{abstract}
\input{sections/abstract.tex}
\end{abstract}

\section{Introduction}\label{sec:introduction}
\input{sections/introduction}

\section{Related work}\label{sec:related}
\input{sections/related}

\section{Knowledge Retention Analysis for Fine-Tuned LLMs}\label{sec:experiment}
\input{sections/experiment}

\section{Model Elasticity Theory}\label{sec:theory}
\input{sections/theory}

\section{Improving Knowledge Retention in Fine-Tuned Models}\label{sec:methods}
\input{sections/methods}

\section{Conclusion and Future Work}\label{sec:Conclusion and Future Work}
\input{sections/conclusion}


\section{Limitations}
While our layer-freezing approach provides a solution for preserving single-edited knowledge, it introduces key limitations. Firstly, it restricts the model's ability to acquire new knowledge during fine-tuning. Secondly, this method would decrease the training efficiency. Lastly, it cannot handle the extreme case that multiple edits that all layers become occupied by prior edits. As a result, development of a method for preserving multiple edits knowledge without compromising model plasticity or training efficiency would be an important future direction. 

\bibliography{main}

\clearpage 

\appendix
\input{sections/supplementary_material}

\end{document}

%% file: sections/abstract.tex
Large language models (LLMs) store vast amounts of knowledge, which often requires updates to correct factual errors, incorporate newly acquired information, or adapt model behavior. Model editing methods have emerged as efficient solutions for such updates, offering localized and precise knowledge modification at significantly lower computational cost than continual training. In parallel, LLMs are frequently fine-tuned for a wide range of downstream tasks. However, the effect of fine-tuning on previously edited knowledge remains poorly understood. In this work, we systematically investigate how different fine-tuning objectives interact with various model editing techniques. Our findings show that edited knowledge is substantially more susceptible to forgetting during fine-tuning than intrinsic knowledge acquired through pre-training. This analysis highlights a key limitation of current editing approaches and suggests that evaluating edit robustness under downstream fine-tuning is critical for their practical deployment. We further find that knowledge retention can be significantly improved by either augmenting edit knowledge with paraphrases or by freezing layers associated with edited content in fine-tuning stage, offering insight for developing more robust editing algorithms.

%% file: sections/introduction.tex
A key factor in the success of large language models (LLMs) is their ability to store vast amounts of knowledge~\citep{radford2019a}. 
The stored knowledge serves as a foundation that enables LLMs to be readily adapted to specific downstream tasks or aligned with human intent through fine-tuning~\citep{Ouyang2022follow}.
Fine-tuning has become an essential step in LLM development, with the majority of production models undergoing this process as developers adapt base models for various applications.
However, the mechanisms regarding knowledge updates during fine-tuning remain unclear, and the risk of catastrophic forgetting persists~\citep{Lange2022,Wu2022,Wang2023}.
As fine-tuning becomes increasingly sophisticated and necessary to LLM development, understanding and improving fine-tuning \textit{knowledge retention} has emerged as a critical challenge.

While LLMs acquire most of their knowledge through pre-training on large corpora, they can also be updated through direct \textit{knowledge editing} (KE).
The KE procedure is analogous to software maintenance: as traditional software requires bug fixes to maintain functionality, LLMs need mechanisms to update their knowledge when it is incorrect or becomes outdated.
For example, when a country elects a new president, the LLM's knowledge regarding the country-president relationship needs to be updated accordingly.
Though such updates could be implemented through continued training, this approach is both costly and risks overfitting~\cite{Mitchell2022SERAC,Gangadhar2024ACL}.
KE methods provide more data-efficient and precise solutions for implementing these updates without full model retraining~\citep{Fang2024,Meng2023MEMIT,meng2023rome,Mitchell2022SERAC}.
However, a crucial question remains: does \textit{edited knowledge} retain during subsequent fine-tuning stages as effectively as \textit{intrinsic knowledge} learned from pre-training?
The retention of edited knowledge through fine-tuning is crucial for LLM development, analogous to ensuring that fixed software bugs do not reappear in later development cycles.



\begin{figure}[t]
    \centering
    \includegraphics[width=1\columnwidth]{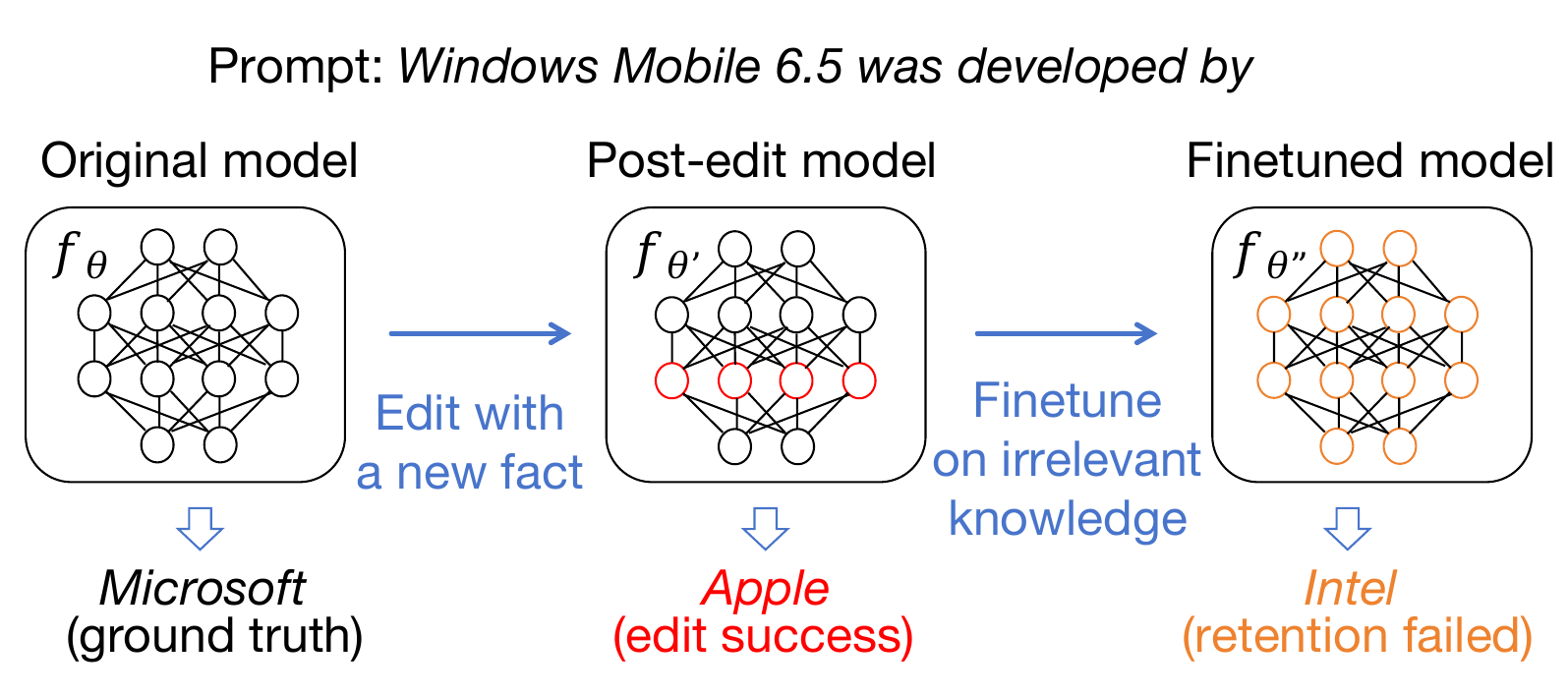} 
    \caption{Demonstration of model editing and downstream model fine-tuning and their impact on the knowledge in LLMs. The original model is edited with a single instance of new fact: \textit{Windows Mobile 6.5 was developed by Apple}, and the edited model is fine-tuned by an irrelevant dataset, which does not contain subject, relation and object from the edited knowledge. Although the edit can be successful, it is vulnerable to different downstream fine-tuning tasks. $f_{\theta}$, $f_{\theta'}$, $f_{\theta"}$ denote the pre-trained models, edited model and fine-tuned model respectively.} 
    \label{fig:figure_intro} 
\end{figure}

In this paper, we start by conducting systematic studies on applying different KE methods on LLMs and evaluating the knowledge retention after fine-tuning, as illustrated in Figure~\ref{fig:figure_intro}. 
We consider four different types of fine-tuning tasks: next token prediction on general text, object prediction for factual knowledge triplets, sentiment classification, and instruction fine-tuning. 
We find that while the knowledge retention rate depends on both the editing method and the task, it is in general \textbf{highly vulnerable} to fine-tuning, and the retention rate of the edited knowledge is significantly lower than that of intrinsic knowledge.

We then connect the knowledge retention with the elasticity theory of LLM fine-tuning~\citep{Ji2025} to explain this fragility.
This theory suggests that the LLM's resistance to deviate from its original distribution is proportional to the training data volume. According to this theory, since the intrinsic knowledge is trained on vastly more data, it achieves greater resistance against fine-tuning than edited knowledge. Therefore, to bridge the gap of retention edited knowledge, we propose two strategies to enhance the resilience of edited knowledge. One is \textbf{editing via augmenting paraphrases}. Experimental results confirm that this data augmentation significantly improves retention, matching or even surpassing the retention rate of intrinsic knowledge when sufficient paraphrases are provided. The other is \textbf{fine-tuning via freezing layers}. We show that selectively freezing layers most associated with the edit content during downstream training effectively preserves edits, achieving retention rates on par with intrinsic knowledge and confirming the localizability of edited knowledge.

The primary contributions of this work are summarized as follows:

\begin{itemize}[leftmargin=0.5em, itemindent=1em]

\item To the best of our knowledge, we conduct the first systematic study on the retention of edited knowledge in LLMs after fine-tuning. We demonstrate that successfully edited knowledge remains highly vulnerable to fine-tuning across a range of editing methods and fine-tuning tasks.

\item We connect the knowledge retention with model elasticity theory to explain the low retention rate of the edited knowledge. In the light shed by this theory, we propose and validate two effective strategies for improving edit resilience. By paraphrase-augmented editing, we expand edit instances with multiple paraphrases of the target fact. By fine-tuning via freezing layers, we selectively freezing layers most associated with the edit content during downstream training. 
\end{itemize}

%% file: sections/related.tex

KE has emerged as a promising paradigm for updating LLMs to adapt to dynamically evolving information. Vanilla fine-tuning is a straightforward method to modify the knowledge in LLMs ~\cite{Zhu2020FT}. There are plenty of works about KE methods, which can be classified into 3 main categories: 1. Locate-then-edit 2. Meta-learning 3. Memory-based.


\textbf{Locate-then-edit.} 
The locate-and-edit method for LLMs knowledge editing involves first identifying the specific parts of the model where the target knowledge is stored ~\citep{Dai2022,Meng2023MEMIT,meng2023rome}. Once localized, the method directly modifies the model’s weights or representations in those areas to update or correct the knowledge. This approach aims to precisely edit knowledge while minimizing broader impacts on the model’s overall performance. ROME and MEMIT are examplars of this category. ROME edits factual knowledge in LLMs by identifying and updating specific rank-one subspaces in the model's weights, allowing precise, localized changes without retraining ~\citep{meng2023rome}. It leverages causal tracing to locate key layers and modifies them efficiently to correct or update facts while preserving the model's overall performance. Building on ROME, MEMIT is a method for mass-editing factual associations in LLMs by directly updating the weights of specific transformer MLP layers identified as causal mediators for factual recall. It spreads the desired memory updates across multiple layers.

\textbf{Meta-learning for KE.} 
Meta-learning for LLMs knowledge editing involves training a hyper-network to generate targeted parameter shifts that update the model's knowledge without full retraining ~\citep{Mitchell2022MEND,Cao2022knowledge}. This approach leverages the hyper-network to transform standard fine-tuning gradients into precise edits, ensuring generalization to semantically equivalent inputs while preserving unrelated knowledge. MALMEN is a representative of the meta-learning-based editing approach. MALMEN edits large language models by using a hyper-network to compute parameter shifts as a least squares problem, solved via the normal equation, enabling efficient and scalable updates while minimizing interference with unrelated knowledge.




Some work has show that KE methods fail to retain the model's accuracy on irrelevant knowledge and general ability \cite{Gupta2024,Gu2024}, and the evaluation paradigm for model editing has been investigated ~\citep{Cohen2024,Zhong2023}. Besides, recent work reveals significant limitations in current methods, showing that their performance on real-world hallucinations often falls short of expectations, and highlights the need for further improvements in the field ~\citep{huang2024}. 

\textbf{Model Elasticity Theory.} 
Recent work has begun to theorize the underlying mechanisms of alignment fragility. \citep{Ji2025} propose the "elasticity" theory, arguing from a data compression ~\citep{Deletang2023} perspective that language models inherently resist distributional shifts from alignment fine-tuning due to the overwhelming influence of pre-training data, causing them to easily revert to pre-trained behaviors.

%% file: sections/experiment.tex
We investigate the retention of both edited and intrinsic knowledge in LLMs after fine-tuning. We evaluate a range of KE methods and fine-tuning tasks under the pipeline: beginning with a base model possessing intrinsic knowledge, we apply a KE technique to introduce edited facts, perform fine-tuning, and subsequently assess the retention of both knowledge types. Here, "knowledge" refers to the model's ability to correctly answer queries regarding a given subject-relation-object triple. They are in the form of $t^c=(s,r,o^c/o^*)$, where s stands for subject, r stands for relation, $o^c$ stands for correct object and $o^*$ stands for false object. The subject and relation together form a prompt used to query the model. It can be evaluated by giving the model a prompt of (subject, relation) and checking if the model can generate the correct object.

Each instance in COUNTERFACT, which is derived from PARAREL, In this work, KE methods are applied to modify the model's association for a given $(s,r)$ pair, replacing the original object $o^c$ with the new, edited object $o^*$ pair.

\subsection{Experiment Setup}
\textbf{KE Dataset:} 
We constructed a dataset to evaluate the retention of both edited and intrinsic knowledge after fine-tuning. Both types of knowledge were derived from the model's existing knowledge base. We use the COUNTERFACT dataset~\cite{Lambert2024tulu}
as the base dataset because it contains factual knowledge in the form of (subject, relation, object) triples as well as the corresponding counterfactual object. We treat the counterfactual object as the edited knowledge, and the factual object as the intrinsic knowledge.
For clean evaluation, we further filter the COUNTERFACT dataset and retain only instances where, given a prompt $p=(s, r)$, the model assigned the highest probability to the true target token by a substantial margin. Furthermore, we ensured that there was no overlap between the intrinsic knowledge dataset and the edit knowledge dataset in terms of subjects, relations, and objects. Further details regarding the dataset are provided in Appendix~\ref{sec:appendix_b}.

\textbf{KE Methods:} We investigate five distinct KE methods:  A baseline method performing full fine-tuning for editing(FT), ROME, MEMIT, MALMEN and AlphaEdit~\cite{meng2023rome,Meng2023MEMIT,Tan2024MALMEN,Fang2024}. For single-edit method (e.g., ROME), the model is updated with one knowledge tuple at a time. For the batch-edit method (e.g., MEMIT), the knowledge dataset is partitioned into groups containing multiple tuples, and the model is updated with an entire group simultaneously to produce the post-edit model. 
For FT and ROME, we also try different edited layer. We find that for both of the methods, edit early layer (prior to layer 10) can achieve similar result.
More details about KE methods are shown in Appendix~\ref{sec:appendix_c1}.

\textbf{Fine-Tuning Tasks and Datasets:} We fine-tune the post-edit model on four different downstream tasks: 1) Next token prediction on general text in CommonCrawl, 2) Object prediction for factual knowledge triplets in COUNTERFACT, 3) Sentiment classification in IMDB, and 4) Instruction fine-tuning in Tulu-3 SFT~\cite{Lambert2024tulu}. 
To prevent potential conflicts between the fine-tuning data and edited/intrinsic knowledge—where the former may contain contradictory information that could degrade the edited knowledge—we systematically filter the fine-tuning dataset. 
To ensure fairness of experiment, we set the same stopping criteria for fine-tuning different post-edit models. 
More details about the fine-tuning tasks are shown in Appendix~\ref{sec:appendix_c2}. 

\begin{figure*}[ht]
    \centering
    \includegraphics[width=1\textwidth,height=0.4\textwidth]{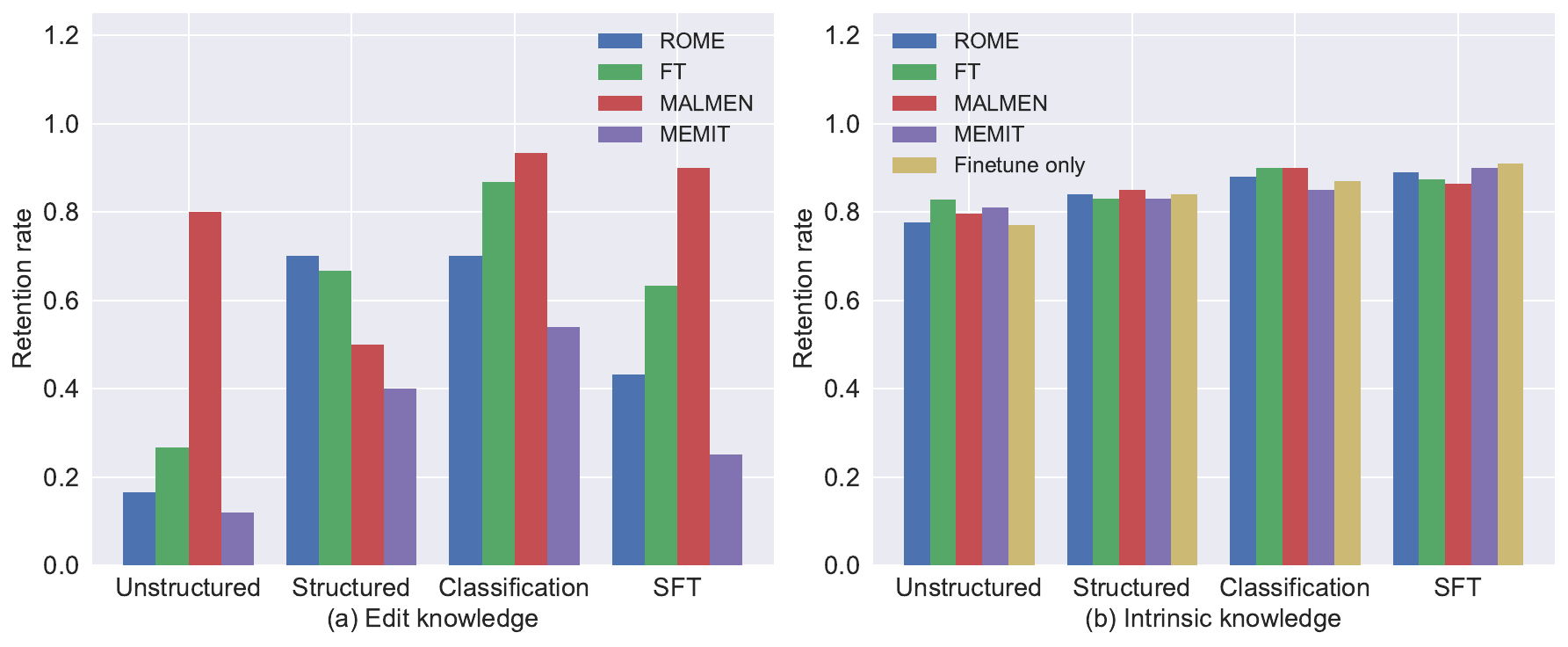} 
    \caption{Edited and intrinsic knowledge retention rate after model edit and fine-tuning for different combination of upstream edit methods and downstream fine-tuning methods. For ROME method, we choose layer 6 as the edit layer. For FT method, we choose layer 1 as the edit layer.}
    \label{fig:figure2} 
\end{figure*}

\textbf{Models:} We perform the KE and fine-tuning on the GPT-2 XL~\cite{radford2019a} and Llama3-8B~\citep{Dubey2024}.

\textbf{Evaluation:} We evaluate the edit and intrinsic knowledge retention rate for both the post-edit model and the fine-tuned model.
To assess whether the model possesses the target knowledge, we input the prompt into the model and compute the probability distribution over the output token. We consider the model to retain the knowledge if the first non-stop token matches the target token. We define the \textit{knowledge retention rate} as the proportion of prompts for which the target token is the highest-ranked non-stop word:
\begin{equation}
\mathbb{E}_{x, y \sim \{(x, y)\}} 1 \left\{ \arg\max_y f_{\theta}(y \mid x) = y_{t},y\notin S\right\} 
\end{equation}

Where $y_{t}$ is the target token, $\{(x, y)\}$ is our edited/intrinsic dataset, S is the set of stopping word first token. We can obtain a intrinsic knowledge retention rate similar to this. We evaluate both the edited knowledge retention rate and intrinsic knowledge retention rate on both post-edit and fine-tuned model. Details about these model edit and fine-tuning hyper-parameters are shown in Appendix~\ref{sec:appendix_c2}.

\subsection{Knowledge Retention After Fine-tuning}

Figure 2(a) and (b) show edited and intrinsic knowledge retention rates after knowledge edit and different fine-tuning tasks on the model of GPT-2 XL. Experiments on a Llama3-8B show similar pattern , as shown in Table~\ref{table:llama}.
\label{sec:appendix_e}

\begin{table}[h]
    \caption{Edit and intrinsic knowledge retention rate after different edit methods and SFT.}
    \centering
    \scalebox{0.95}{
    \begin{tabular}{llll}
    \toprule
    Edit method & ROME & FT & MALMEN \\
    \midrule
    Edited knowledge & 0.374 & 0.634 & 0.815 \\
    Intrinsic knowledge & 0.824 & 0.831 & 0.817  \\
    \bottomrule
    \end{tabular}
    }
    \label{table:llama}
\end{table}

To evaluate knowledge retention in a larger-scale model and a different model architecture, we conduct SFT experiments on Llama3-8B. We employ the same methodology as GPT-2 XL, but with a larger training dataset and stricter stopping criteria. We observe similar result as GPT-2 XL: Edited and intrinsic knowledge has similar retention rate for MALMEN. However, for FT and ROME, edited knowledge retention rate is much lower than intrinsic knowledge. 

Notably, while MALMEN maintains comparable retention rates for edited and intrinsic knowledge, this does not imply that edited knowledge is the inherently same as intrinsic knowledge. The intrinsic knowledge retention rate has a drop during model edition, from 1.0 to ~0.8, whereas the edited knowledge retention rate starts at 1.0 post-editing.

\textbf{The edited knowledge exhibit lower retention rate than the intrinsic knowledge, which suggests that even when a fact is successfully inserted into a model, it is still inherently different from the intrinsic knowledge.} 
MALMEN achieves the best overall performance: it presents similar edit and intrinsic knowledge retention rate for unstructured data fine-tuning, classification, and SFT task, but lower retention rate for structured fine-tuning task. ROME/MEMIT demonstrates lowest knowledge retention result for unstructured, classification and SFT task, but best for the structured task. This does not contradict to the model elasticity theory, as different KE methods have different effectiveness, thus different compression rate. 

\textbf{In contrast to edited knowledge, after fine-tuning, the intrinsic knowledge retention rate remains nearly the same for all of these tasks.} This matches with the conclusion of model elasticity theory. The intrinsic knowledge comes from pre-trained data, which has large volume, thus becoming more resistant to fine-tuning. 


\subsection{Evolvement of Token Distribution During Fine-Tuning} 
We further analyze the distribution of output token during downstream fine-tuning after model editing. We show the four categories of output tokens in the Figure~\ref{fig:figure3}.

\newcommand{\lminput}[1]{``\textsf{#1}''}
\newcommand{\problembody}[2]{%
  \begin{tabular}{|p{\dimexpr0.5\textwidth-2\tabcolsep\relax}|}
    \hline
    \textcolor{white}{\textbf{\large #1}} \\
    \hline
    \begin{minipage}{\dimexpr0.5\textwidth-2\tabcolsep-2\fboxsep\relax}
      \vspace{0.5em}
      #2
      \vspace{0.5em}
    \end{minipage} \\
    \hline
  \end{tabular}
}
\newcommand{\boxline}{%
  \vspace{1em}
  \hrule
  \vspace{1em}
}

\begin{figure}[t]  
  \centering
  \problembody{Truthful Intervention Case Study}{
    Token type: True token
    
    Definition: The original correct token in $t^{c}$
    
    Example: \textit{Microsoft}
    \boxline

    Token type: Edited target token
    
    Definition: The target editing token in $t^{*}$
    
    Example: \textit{Apple}
    \boxline
    
     Token type: Related-term token
     
     Definition: 
     Tokens belong to the same category. The sentence's meaning would be changed if substituted
     
    Example: \textit{Intel, IBM, Google}
     \boxline
     
     Token type: Other tokens
     
     Definition: Completely unrelated tokens, but the sentence would most likely be grammatically correct 
     
     Example: \textit{astronaut, Egypt}
  }
  \caption{Different type of output tokens for the prompt "\textit{Windows Mobile 6.5 was developed by}" as an example.}
  \label{fig:figure3}
\end{figure}

\begin{figure*}[t]
    \centering
    \includegraphics[width=1\textwidth,height=0.4\textwidth]{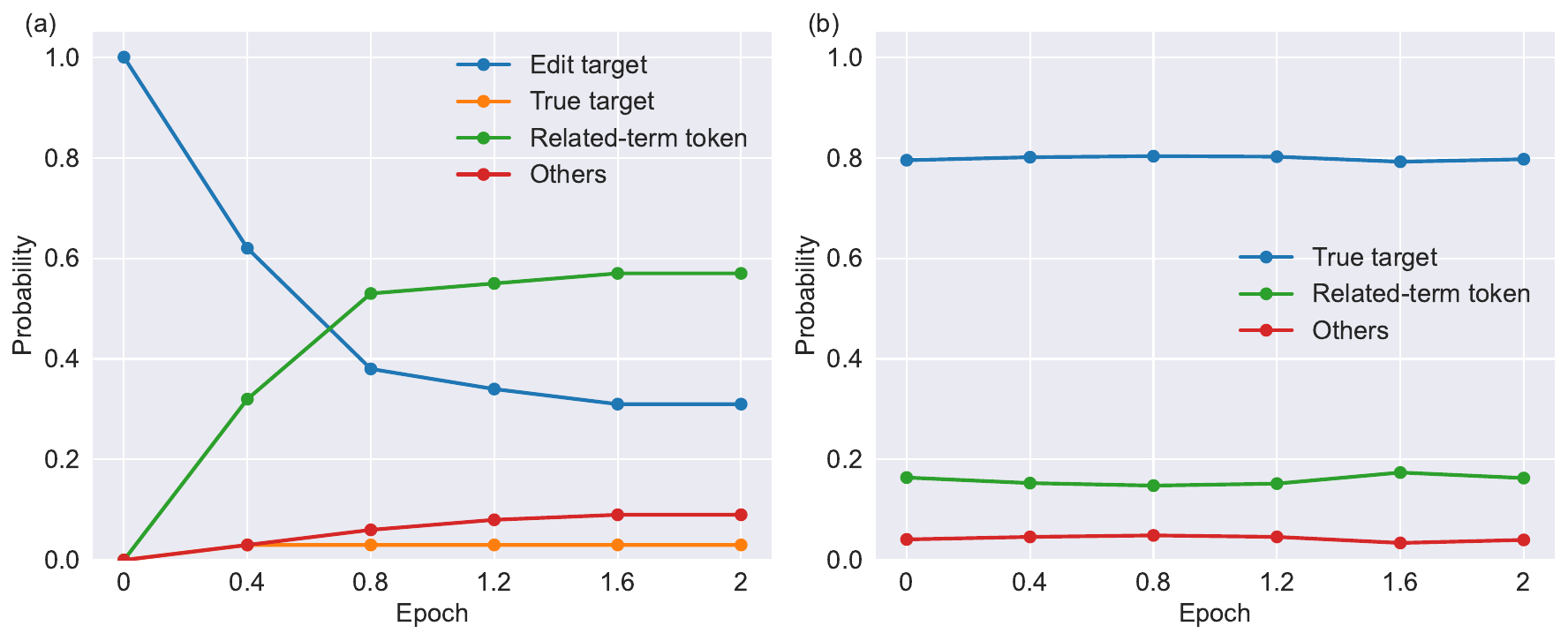} 
    \caption{(a) First generated token distribution vs training epoch for edited knowledge. (b) First generated token distribution vs training epoch for intrinsic knowledge. Initially, the model predict the target token of "\textit{Microsoft}" correctly. After model edition, the post-edit model predict an extremely high probability of 0.992 to the edited target token "\textit{Apple}", and the true target Microsoft has very low probability(lower than 0.001). However, after just one and a half epoch of fine-tuning, this token disappears from the top-3 predicted tokens. The top-3 token rankings stabilize after one epoch of fine-tuning.} 
    \label{fig:figure4} 
\end{figure*}

Figure~\ref{fig:figure4} illustrates the evolution of output token distributions throughout the fine-tuning process for both edit knowledge and intrinsic knowledge, when given prompt. We conduct experiments on ROME as the editing method and using unstructured datasets for fine-tuning. We average the probability of the first output token of all the 50 experiments.

We see that the probability of generating edit token first drops quickly, and the probability of generating related-term token increases quickly, and both of them finally converge. \textbf{Our results show that after fine-tuning the post-edit model, the model would unlikely generate the original true target, but the related-term token.} Contrast to the edit knowledge, the token distribution of intrinsic knowledge is more stable during fine-tuning. 

\input{tables/top_token}

We present a case study using a knowledge tuple chosen from COUNTERFACT in Table~\ref{table:table1}. The knowledge is edited using ROME, followed by fine-tuning the post-edit model on unstructured datasets. 
Interestingly, while the true target \textit{"Microsoft"} does not achieve high probability, "\textit{Intel}"—a related-term token—shows increasing probability. Notably, even after removing all the text that contains non-stop keyword tokens from the prompt (\textit{"window"}, \textit{"mobile"}, and \textit{"develop"}) in the fine-tuning dataset, this related-term tokens can still achieve the highest ranking after fine-tuning.

%% file: tables/top_token.tex
\begin{table*}[t]
\caption{Model's Top-3 tokens and their probability in different stages for the prompt of "\textit{Windows Mobile 6.5 was developed by}", true target of "\textit{Microsoft}" and the edit target of "\textit{Apple}".}
\centering
\begin{tabular}{>{\centering\arraybackslash}m{3cm}|>{\centering\arraybackslash}m{8cm}}
\toprule
Model & Top3-tokens \\ \midrule
Original model & Microsoft (0.258) | Nokia (0.201) | the (0.107) \\ \midrule
Post-edit model & Apple (0.992) | Nokia (0.004) | Google (0.001) \\ \midrule
0.4 finetuned epoch & Apple (0.297) | Intel (0.179) | Nokia (0.004) \\ \midrule
0.8 finetuned epoch & Intel (0.183) | Apple (0.126) | IBM (0.056) \\ \midrule
1.2 finetuned epoch & Intel (0.170) | Nokia (0.051) | Apple (0.049) \\ \midrule
1.6 finetuned epoch & Intel (0.254) | Nokia (0.210) | Google (0.106) \\ \midrule
2 finetuned epoch & Intel (0.247) | Nokia (0.147) | Google (0.08) \\ \bottomrule
\end{tabular}
\vspace{1em} 
\label{table:table1} 
\end{table*}

%% file: sections/theory.tex
The elasticity theory of LLMs provides a framework to understand the behavior of a fine-tuned model that is pre-trained with a mixture of different datasets~\cite{Ji2025}. The theory can be represented by Equation \ref{eq:elasticity}, when the model is firstly trained with $\mathcal{D}_{1}$, $\mathcal{D}_{2}$, and subsequently trained with $\mathcal{D}_{3}$:

\begin{equation} \label{eq:elasticity}
\frac{d\gamma_{p_{\theta}}^{\mathcal{D}_{2}/\mathcal{D}}}{d\,l} =\Theta\left(-k\frac{d\gamma_{p_{\theta}}^{\mathcal{D}_{1}/\mathcal{D}}}{d\,l}\right)
\end{equation}

where $l = \frac{\left|\mathcal{D}_3\right|}{\left|\mathcal{D}_2\right|} \ll 1$,
$k = \frac{\left|\mathcal{D}_1\right|}{\left|\mathcal{D}_2\right|} \gg 1$,

and $\gamma$ is the normalized compression rate.

A lower normalized compression rate corresponds to reduced training loss on a given piece of knowledge, indicating that the model has mastered the knowledge more thoroughly, which in turn leads to a higher retention rate.

In our setting of studying the knowledge retention in fine-tuning, $\mathcal{D}_{1}$ corresponds to the pre-training dataset for a given piece of knowledge, $\mathcal{D}_{2}$ corresponds to the dataset used for editing, and $\mathcal{D}_{3}$ corresponds to the dataset used for the subsequent fine-tuning. A key distinction between edited knowledge and intrinsic knowledge lies in the diversity of their training expressions: intrinsic knowledge is acquired from a wide variety of paraphrases present in the pre-training corpus, whereas edited knowledge is typically introduced through a single formulation.
For example, during pre-training, an LLM may encounter a fact such as "\textit{Windows Mobile 6.5 was developed by Microsoft}" in various paraphrased forms—such as "\textit{Windows Mobile 6.5 is a product of Microsoft}" or "\textit{Windows Mobile 6.5 was created by Microsoft}." According to model elasticity theory, this multiplicity of expressions helps consolidate the internal representations of the knowledge, thereby enhancing its retention to subsequent fine-tuning.

%% file: sections/methods.tex


To enhance the resilience of edited knowledge against downstream fine-tuning, we investigate two complementary strategies: one that addresses the fundamental cause during the model editing stage, and another that serves as an efficient adaptation during the fine-tuning stage.

\subsection{Edit with Augmented Paraphrases} 
\begin{figure*}[ht]
    \centering
    \includegraphics[width=1\textwidth,height=0.4\textwidth]{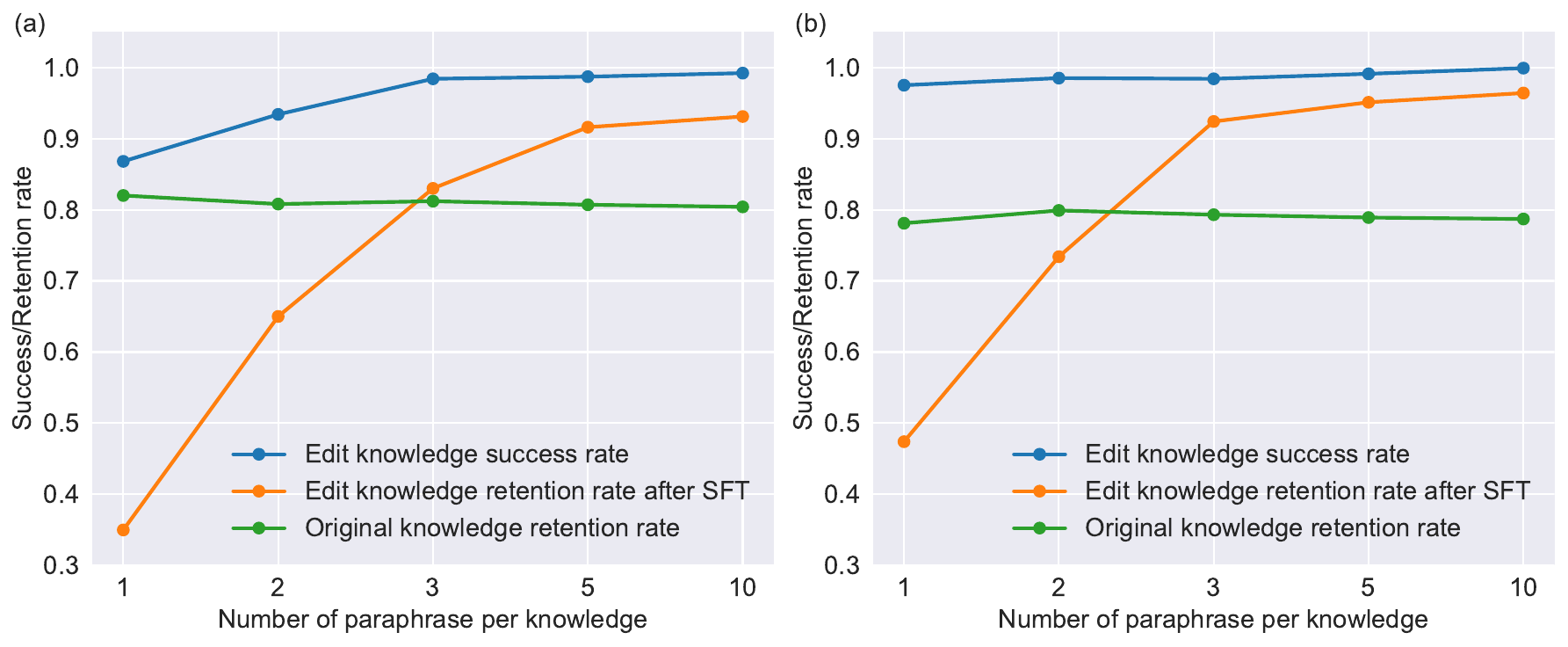} 
    \caption{Success/Retention rate vs number of paraphrases per knowledge fact for the method of (a) MEMIT (b) AlphaEdit} 
    \label{fig:figure5} 
\end{figure*}

Based on the insight from model elasticity theory, the retention of edited knowledge against subsequent fine-tuning is positively correlated with the volume of data used during the editing phase. We hypothesize that, to achieve a post-SFT edited knowledge retention rate comparable to that of intrinsic knowledge, each edit knowledge should be edited with comparable number of paraphrases for intrinsic knowledge used during pre-training. Moreover, if the number of paraphrases provided during editing exceeds the average lexical variety encountered for facts in the pre-training corpus, the retention of edited knowledge may even exceed that of intrinsic knowledge. 

To test this, we augmented the edit dataset by generating multiple paraphrases for each knowledge fact and evaluated their impact on retention rates after SFT. In this framework, augmenting the editing paraphrases plays the same role with increasing the effective training data volume for that knowledge.

We evaluated this approach using batch-editing methods, including MEMIT and AlphaEdit. Each method was applied to edit a batch of 100 facts, with each fact associated with [1, 2, 3, 5, 10] paraphrased statements. We subsequently compared the retention rates of both edited and intrinsic knowledge following SFT on the Tulu-3-SFT-mixture dataset. The experimental results are summarized in Figure \ref{fig:figure5}. Further details regarding model editing and fine-tuning hyperparameters are provided in Appendix~\ref{sec:appendix_b}.

The editing success rate and the retention of edited knowledge both improve with an increasing number of paraphrases per fact. In contrast, the retention rate of intrinsic knowledge remains stable at approximately $80\%$. Notably, beyond a certain threshold of paraphrase,  the retention of edit knowledge surpasses that of intrinsic knowledge. We posit that this occurs because, during pre-training, each piece of intrinsic knowledge fact is acquired from a limited number of paraphrases. When the number of paraphrases for an edited fact meets or exceeds the average number of knowledge variants encountered during pre-training, its retention resilience becomes comparable to, or even greater than, that of intrinsic knowledge. Our experiments indicate that providing approximately 3 paraphrases is sufficient for edit knowledge to match the retention rate of intrinsic knowledge.

\subsection{Fine-Tuning with Frozen Selected Layers} 

For Locate-then-Edit method, the layer that the edited knowledge being inserted into is where the knowledge located ~\cite{meng2023rome,Geva2021keyvaluememory}. We hypothesize that preserving edited knowledge can be improved by avoiding fine-tuning of layers containing the target knowledge. To test this hypothesis, we evaluate two layer-specific fine-tuning strategies on GPT-2 XL:

\begin{figure*}[ht]
    \centering
    \includegraphics[width=1\textwidth,height=0.4\textwidth]{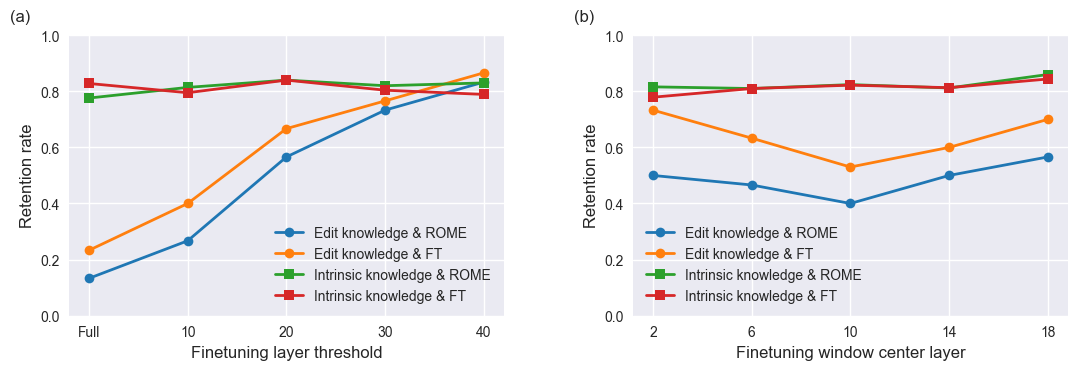} 
    \caption{(a) Edit and intrinsic knowledge retention rate for different finetuning layer threshold for Rome and FT edit.  We choose layer 1 as the edit layer for both ROME and FT method. (b) Edit and intrinsic knowledge retention rate for different finetuning window center layer for ROME and FT edit. We choose layer 10 as the edit layer for both ROME and FT method.} 
    \label{fig:figure6} 
\end{figure*}

\textbf{Finetune only later layers.} We freeze the early layers and fine-tune the layers beyond a specified threshold layer. As GPT-2 XL contains 48 transformer layers, we set the finetuning layer threshold to be 10, 20, 30, 40 and compare their result. From the Figure~\ref{fig:figure6}(a), we find that larger layer threshold can improve the edited knowledge retention rate while having similar intrinsic knowledge retention rate. If only layers after 40 are finetuned, the edited knowledge reaches the same level of knowledge retention with intrinsic knowledge. 
In addition, finetuning after a layer threshold has similar intrinsic knowledge retention rate with full finetuning retention rate, showing that this method does not impair the intrinsic knowledge. 

\textbf{Finetune only layers in a window.} We freeze all layers except a window of layers. We focus on two methods: ROME and FT, because both of them incorporate knowledge by editing a single layer. We set the window size to be 5. To reserve a larger room of layers for the experiment, for both FT and ROME, we choose layer 10 instead of 6. From the Figure~\ref{fig:figure6}(b), we observe that fine-tuning achieves lowest edited knowledge retention rate when the window centered at layer 10-where the edited knowledge is located-while exhibiting higher edited knowledge retention rate for window does not include the edit layer, or even centered farther from this layer. In contrast, intrinsic knowledge maintains a similar knowledge retention rate across different window center positions.

%% file: sections/conclusion.tex
We examine how different downstream fine-tuning tasks affect previous edited knowledge. First, we demonstrate that knowledge incorporated via KE methods are particularly sensitive to downstream fine-tuning data, and none of ROME, MEMIT, FT and MALMEN can edit the knowledge to retain as intrinsic knowledge that suit for all the downstream fine-tuning tasks: they usually output related tokens that belong to the same category rather than the true target token. 

Finally, we propose two solutions to improve the resilience of edited knowledge, the first approach increases the number of paraphrased expressions during the editing phase, which fundamentally make the edited knowledge comparable to intrinsic knowledge, the second implements a practical fine-tuning adaptation through selective layer freezing, preserving edits by avoiding modifications to layers most critical to the edited content.

For future KE methods, post-edit retention should be assessed not only immediately after editing, but also after fine-tuning. An interesting research direction involves developing editing techniques that automatically instill knowledge with a high retention rate, potentially leveraging models' paraphrasing capabilities to create edits inherently resistant to fine-tuning drift.

%% file: sections/supplementary_material.tex
\section{Additional Related Work} 
\label{sec:appendix_a}
Memory-based methods store edits in a explicit memory without modifying the model's parameters ~\citep{Mitchell2022SERAC,Zheng2023ACL,Hartvigsen2023}. For example, SERAC is a gradient-free memory-based model editing method that stores edits in an explicit memory and uses a scope classifier to determine if a test input is within the scope of any cached edits. If within scope, a counterfactual model predicts the label based on the most relevant edit example; otherwise, the base model's prediction is used ~\citep{Mitchell2022SERAC}.

\section{Dataset Details} 
\label{sec:appendix_b}
\begin{table*}[thbp]
    \caption{Information for fine-tuning dataset.}
    \centering
    \scalebox{0.95}{
    \begin{tabular}{llll}
    \toprule
    Task name & Data source & \#Train data points & \#Validation data points \\
    \midrule
    Unstructured & Common Crawl & 60,000 & 1,000 \\
    Structured & CounterFact & 3,000 & 100  \\
    Classification &  IMDB & 25,000 & 100  \\
    SFT (GPT-2 XL) & Tulu-3-SFT-mixture dataset & 10,000  & 1,000 \\
    SFT (Llama3-8B) & Tulu-3-SFT-mixture dataset & 100,000  & 1,000 \\
    \bottomrule
    \end{tabular}
    }
    \label{tab:datasets}
\end{table*}
In CounterFact, given prompt, we observe that models can generate top-ranked tokens with nearly identical probabilities(difference less than 0.03), indicating the model is uncertain about the knowledge even it may predict the true target correctly. To mitigate this ambiguity and ensure the model confidently possesses the target knowledge, we filter both the edit and intrinsic knowledge datasets, retaining only samples where the true target token’s probability exceeds the second-ranked token’s probability by at least 0.1. 

Applying this criterion, we filter out 1,434 and 4,939 samples from the CounterFact training split for GPT-2 XL and Llama3-8B, respectively. From the remaining data, we sample 50 prompts for the edit dataset and 100 prompts for the intrinsic dataset. Detailed statistics of the filtered datasets are provided in Table 3.
\begin{table*}[thbp]
    \caption{Information for edit and intrinsic data for GPT-2 XL and Llama3-8B.}
    \centering
    \scalebox{0.95}{
    \begin{tabular}{llll}
    \toprule
    Task name & \# Data points & Avg true target prob & Avg second ranked target prob  \\
    \midrule
    GPT-2 XL Edit & 50 & 0.407 & 0.088 \\
    GPT-2 XL Intrinsic & 100  & 0.409 & 0.092 \\
    Llama3-8B Edit & 50 & 0.467 & 0.112 \\
    Llama3-8B Intrinsic & 100  & 0.473 & 0.119 \\
    \bottomrule
    \end{tabular}
    }
    \label{tab:datasets1}
\end{table*}

As the GPT-2 XL and Llama3-8B possess different knowledge on the CounterFact dataset, necessitating the construction of different edit datasets and intrinsic knowledge datasets for each model. Compared to GPT-2 XL, Llama3-8B not only demonstrates more comprehensive knowledge coverage but also exhibits significantly higher prediction confidence for the target facts in CounterFact.

\section{Implementation Details} 

\subsection{Knowledge Editing} 
\label{sec:appendix_c1}
To enhance locality, we fine-tune only a specific layer while freezing all others, as this approach demonstrates superior locality compared to full-model fine-tuning~\citep{Gangadhar2024ACL}. A study has been conducted to determine the optimal layer for knowledge edits ~\citep{Hase2024NIPS,meng2023rome}. For FT and ROME method, we choose layer 1 and layer 6 to be the edit layer respectively. 

\textbf{Full fine-tuning for editing(FT):} 
We choose Adam optimizer with learning rate of 5e-5, maximum training step of 25, weight decay of 0 and early stopping loss of 0.01. We fine-tune one specific layer's $mlp_{proj}$ of the model. We choose layer 1 as the edit layer for both GPT-2 XL and Llama3-8B.

\textbf{ROME:} 
We emply the same hyper-parameters for ROME as the setting in original paper ~\cite{meng2023rome}: We choose learning rate of 0.5, maximum training step of 50, weight decay of 0.5 and KL factor of 0.0625. We perform model edition on one layer of the model. We choose layer 6 as the edit layer for both GPT-2 XL and Llama3-8B.

\textbf{MEMIT:} 
We choose learning rate of 0.2, maximum training step of 50, weight decay of 0.003, editing layers ranging from 3 to 8 and KL factor of 0.0625. We edit 10 facts in each model edition, so we perform 5 model edition experiments for each fine-tuning task.

\textbf{MALMEN:} 
We adopt the same hyper-parameters for MALMEN as those used in the original paper ~\cite{Tan2024MALMEN}.   
We observe that choosing later layers can achieve better edition success rate. We select the model editing hyperparameter such that the edited model can predict true target for all the prompts in our edit dataset. Specifically, for GPT-2 XL, we edit layers ranging from 43 to 48(out of 48), while for Llama3-8B, we edit layers ranging from 27 to 32(out of 32).

\textbf{AlphaEdit:} 
We adopt the same hyper-parameters for AlphaEdit as those used in the original paper ~\cite{Fang2024}.   
We select the model editing hyperparameter such that the edited model can predict true target for all the prompts in our edit dataset. For GPT-2 XL, we edit layers ranging from 13 to 17(out of 48).

\subsection{Downstream Fine-tuning Tasks} 
\label{sec:appendix_c2}

We perform data filtering on the training set of all the downstream task, only keep the data that are irrelevant to the edit knowledge. To fairly compare these edit methods, it is essential to develop a metric to quantify the extent to which fine-tuning influences the model. To ensure that the fine-tuning impact is consistent across the four methods (FT, ROME, MEMIT and MALMEN), for each of these downstream fine-tuning tasks, we sample and fix an evaluation dataset from the fine-tuning dataset and standardize the training stopping criteria, and employ the same fine-tuning hyperparameter.

For a fine-tuning task, ensuring consistent fine-tuning impacts across different post-edit models is challenging but critical for fairness. We set the same stopping criteria for fine-tuning different post-edit models. Some experiments take more training epoch to reach the stopping criteria. For example, we note that fine-tuning with larger layer threshold needs larger training epoch. Notably, experiments with larger layer thresholds require more training epochs to meet these criteria, as fewer parameters are updated, demanding greater training effort to incorporate the same volume of knowledge into the model. For FT and ROME, we also try different edited layer. We find that for both of the methods, edit early layer (prior to layer 10) can achieve similar result.

\textbf{Next token prediction on general text} To systematically assess the impact on both the model’s intrinsic knowledge and the edited knowledge, we employed an unstructured dataset distinct from all pre-training corpora. Specifically, for our experiments on the GPT-2 XL, which was pre-trained on webtext, we choose the Common Crawl dataset for fine-tuning. Since a small subset of the data suffices to demonstrate the influence, we sampled 60k data for our training set. We evaluate the fine-tuning influence on a validation set, which contains of 1k instances sampled from the training set. We employ loss as a metric to measure the volume of knowledge acquired from fine-tuning. The fine-tuning is stopped once the validation loss goes below the threshold of 3.1.

\textbf{Object prediction for factual knowledge triplets} To investigate the influence of structured factual data on model behavior, we construct a fine-tuning dataset that mirrors the structure of our edit data while containing distinct factual content. Specifically, we sample data from COUNTERFACT train split, which has no overlap with the one used in upstream edit. We select only instances where the pre-trained model fails to predict the true target, and sample 3,000 of them. For supervision, we use the true target from each example as the training label. To monitor progress, we evaluate model performance on a validation set, which contains 100 instances sampled from the training set. Training terminates once the model achieves above 70\% accuracy on this validation set.

\textbf{Sentiment classification task}
To assess the impact of downstream fine-tuning on pre-trained knowledge, we employ a classification task as our benchmark evaluation. Specifically, we utilize the IMDB sentiment analysis dataset which consists of 25k movie reviews paired with binary sentiment labels. For the classification architecture, we append a fully connected layer to the final hidden state of the end-of-sequence (EOS) token. Classification accuracy on the validation set can be a fair metric, as it is the target index people are aiming to improve. The validation set contains 100 samples, which has no overlap with the training set. We fine-tune the model until the model achieves a classification accuracy exceeding 95\% on the validation set.

\textbf{Instruction fine-tuning task}
Supervised fine-tuning has emerged as a prevalent downstream adaptation method for LLMs. In this paradigm, each training instance consists of an instruction (question) paired with its corresponding response (answer). For our experiments, we employ the Tulu-3-SFT-mixture dataset ~\cite{Lambert2024tulu} for instruction fine-tuning, which consists of 10k movie reviews paired with binary sentiment labels. To assess the training influence, we evaluate model performance on a validation set, which contains 1k instances sampled from the training set. Training terminates once the model achieves above 70\% accuracy on this validation set. We evaluate the training progress by compute the model's loss on the validation set.  We set the stopping criteria as validation loss goes below the threshold of 2.5.

\subsection{Hyperparameter for fine-tuning} 
\label{sec:appendix_c3}
\textbf{Unstructured fine-tuning:} 
For GPT-2 XL, in align with its pre-training stage, we employ batch size of 256, AdamW optimizer with learning rate of 5e-5, beta of (0.9,0.999) and weight decay of 0.01. The training data set contains 60k data, and the validation data set contains 1k data. The model is evaluated on the validation set every 20 steps. The training stops once the validation loss goes below 3.1.

\textbf{Structured fine-tuning:} 
We employ batch size of 256, AdamW optimizer with learning rate of 2e-5, beta of (0.9,0.999) and weight decay of 0.01. The training data set contains 3,000 data, and the validation data set contains 100 data. The model is evaluated on the validation set every 20 steps. The training stops once the accuracy on the validation set goes above 70\%. 

\textbf{Sentiment classification:} 
We employ batch size of 64, AdamW optimizer with learning rate of 2e-5, beta of (0.9,0.999) and weight decay of 0.01. The training data set contains 25,000 data, and the validation data set contains 100 data. The model is evaluated on the validation set every 5 steps. The training stops once the accuracy on the validation set goes above 95\%. 

\textbf{Supervised fine-tuning(SFT):} 
For GPT-2 XL, we use an AdamW optimizer with learning rate of 2e-5, with beta of (0.9,0.999) and weight decay of 0.01. The training data set contains 10,000 data, and the validation data set contains 1,000 data. Training is performed with a batch size of 64, and the model is evaluated on the validation dataset every 20 steps. We halt training once the validation loss falls below 1.7. 

For Llama3-8B, we employ AdamW optimizer with learning rate of 2e-5, beta of (0.9,0.999), batch size of 256, and weight decay of 0.01. The training data set contains 100,000 data, and the validation data set contains 1,000 data. The model is evaluated on the validation dataset every 20 steps. Due to its larger capacity and greater training requirements, it needs more training effort, so we set the validation loss of 1.0 as the training stopping criteria and evaluate the model on the validation set every 20 steps. 

\subsection{Knowledge Retention Rate} 
\label{sec:appendix_c4}
The model can sometimes assigns the highest probability to a stop word but not the target word. However, this does not necessarily imply a lack of knowledge. In such cases, we employ greedy decoding to generate subsequent tokens until a non-stop token is produced.

\subsection{Experiment Results for Different Edit Layer} 
\label{sec:appendix_d}

\begin{figure*}[h]
    \centering
\includegraphics[width=0.7\textwidth,height=0.4\textwidth]{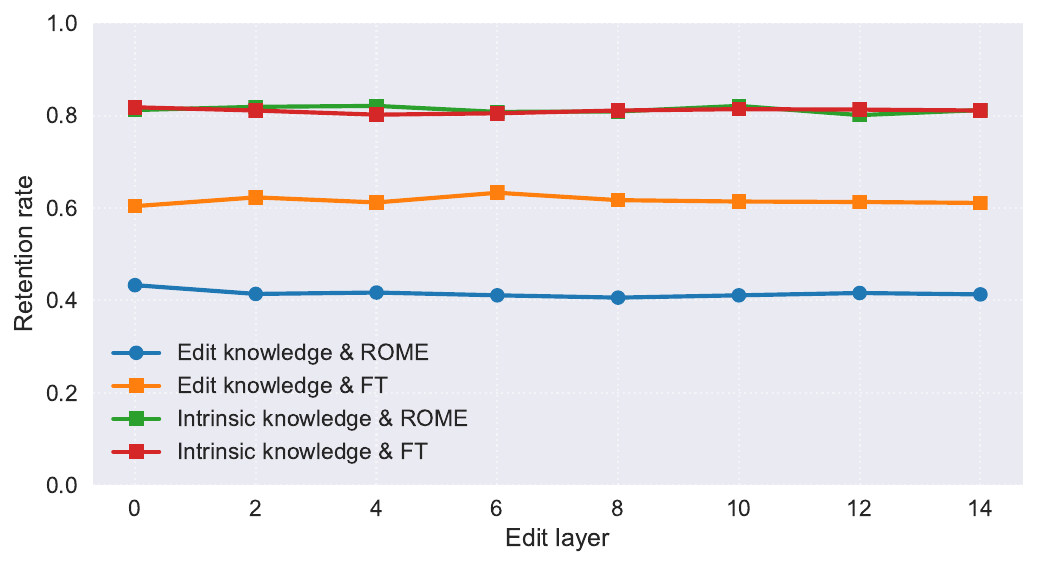} 
    \caption{Edit and intrinsic knowledge retention rate for different edit layer for ROME and FT methods.} 
    \label{fig:figure7} 
\end{figure*}

We analyze how the choice of editing layer influences model performance during downstream SFT on GPT-2 XL. Figure 5 compares knowledge retention rates for both edited and intrinsic knowledge across different layers when applying ROME and fine-tuning (FT) methods. Our experiments demonstrate that for early layers, ROME and FT achieve comparable edited knowledge and intrinsic knowledge retention rates across all edited layers.